\documentclass[10pt,aps,prl,twocolumn,showpacs,superscriptaddress,amsmath,amssymb]{revtex4-1}

\usepackage{amsmath,amssymb,graphicx}
\usepackage{algpseudocode}
\usepackage{bm}
\usepackage{color}
\usepackage{graphicx}
\usepackage{xcolor}
\newcommand\numberthis{\addtocounter{equation}{1}\tag{\theequation}}

\newcommand{\norm}[1]{\left\lVert#1\right\rVert}

\begin{document}
\title{Intensity-only optical compressive imaging using a multiply scattering material and a double phase retrieval approach}

\author{Boshra Rajaei$^{1,5, 6}$, Eric W. Tramel$^{2}$, Sylvain Gigan$^{3,6}$, Florent Krzakala$^{2,6}$, Laurent Daudet$^{1,4,6}$}
\noaffiliation
\affiliation{
Institut Langevin, ESPCI and CNRS UMR 7587, Paris, F-75005, 
France\\
$^2$LPS-ENS, UPMC and CNRS UMR 8550, Paris, F-75005, France.\\
$^3$Laboratoire Kastler Brossel, UPMC, ENS, Coll\`ege de France, CNRS UMR 8552, Paris, F-75005, France \\
$^4$Paris Diderot University, Sorbonne Paris Cit\'e, Paris, F-75013, France\\
$^5$Sadjad University of Technology, Mashhad, Iran\\
$^6$ PSL Research University, F-75005 Paris, France
}

\begin{abstract} In this paper, the problem of
compressive imaging is addressed using natural randomization by means of a
multiply scattering medium. To utilize the medium in this way, its corresponding
transmission matrix must be estimated. For calibration purposes, 
we use a digital micromirror device (DMD) as a simple, cheap,
and high-resolution binary intensity modulator. 
We propose a phase retrieval algorithm  which is well adapted to intensity-only measurements on the camera, and to the input binary intensity patterns, both to estimate the complex transmission matrix as well as image reconstruction. 
We demonstrate promising experimental results for the proposed double phase retrieval algorithm
using the MNIST dataset of handwritten digits as example images.
 \end{abstract} 
\maketitle

\section{Introduction}\label{sec:Introdution}  
From the perspective of image processing, 
the goal of compressed sensing (CS) 
is to reconstruct a high-resolution
image, which is sparse in either the ambient domain or some transform basis,
using few incoherent linear projections \cite{romberg08}. 
Over the past decade,
there has been a tremendous amount of work in the field of CS,
including analytical reconstruction guarantees as well as developments of new
algorithmic approaches that provide efficient methods of solving the 
reconstruction task \cite{eldar12,boche15}. 
However, to date there have been only a handful of
engineering projects where optical imagers based on CS have actually been
built. Indeed, performing these incoherent, usually random, projections is a
highly non-trivial task, requiring innovative hardware solutions.  Amongst
such imagers, one can cite, without any claim of completeness, several 
single-pixel imaging systems \cite{duarte08,magalhaes11,shrekenhamer13}, a
random lens camera \cite{fergus06}, and an imaging setup based on a rotating
diffuser \cite{zhao12}.

The work presented in this paper is built upon a recently developed
optical CS setup \cite{dremeau15} that uses a multiply scattering medium
to effect the random projection operation.
The fundamental difference with this approach and 
most of the CS systems discussed above is that
here the random projections are not designed beforehand and then implemented
through sophisticated hardware, as in \cite{gao14}, but are based on the
natural randomization properties of coherent light multiply scattering
through a layer of opaque material. 
Here, the word ``multiply'' refers to the
fact that the thickness of the material slab is many times larger than 
the mean free path,
ensuring that the light beam is fully scattered without any remaining
ballistic photons at the output. If $x$ is the incoming wavefield (the object
to be imaged at the input plane), the scattering operation is well
modeled by a simple linear operator $\mathbf{H}$, called the transmission matrix. 
If $y$ is the output wavefield discretized by receptor pixels, 
then, in the ideal noiseless case,
\begin{equation}
\label{eq:TM_model}
y = \mathbf{H} x. 
\end{equation}
It has been shown that the transmission matrix of a scattering material is
statistically identical to an i.i.d. random matrix with a complex Gaussian
distribution \cite{popoff10}. The benefits of using such a system for CS imaging
are that one does not have to rely on complex engineering solutions to provide
the (pseudo-) randomization, and also that, in theory, only one shot is
necessary to obtain any desired number of output features; as opposed to
the single-pixel camera which intrinsically requires sequential measurements. 

There exists, however, an obvious price to pay: the necessity of a precise calibration step.
Indeed, to be able to use this system as an imaging device, i.e. to estimate
$x$ given measurements $y$, one must have accurate knowledge of the
matrix $\mathbf{H}$. This can be accomplished by sending a series
of known images, measuring the corresponding outputs, and performing a 
least-squares estimate of $\mathbf{H}$. The calibration step is conducted
by shaping the input wavefront with a Spatial Light Modulator (SLM), 
which is only used for calibration and display, and is not part of the 
direct imaging system.

In this paper, we circumvent one major limitation of the previous 
proof-of-concept system \cite{liutkus14}. Since optical sensors (here, a CCD camera)
only measure the field intensity $|y|^2$, in \cite{liutkus14} the input image is phase-modulated
using a phase-only SLM, with relative phases $0$, $\pi/4$,  $\pi/2$, and $3
\pi/4$. Combining the corresponding four output intensity images, one can
easily recover the complex field $y$ using a method known as ``phase-stepping
holography''. Furthermore, such phase-only modulated images have a constant
intensity. To obtain an image that is sparse in the spatial domain, one has to
make the difference between 2 complex phase-only images which only differs
by a sparse number of pixels. Therefore, in order to get the complex
measurements corresponding to a single sparse image, 8
intensity measurements are required. This significantly slows down both the calibration and
the measurement process. Furthermore, a sufficiently fast continuous-phase SLM
is a very expensive device, with limited pixel counts. For example, the SLM used in
\cite{liutkus14} could only display $32\times32$ images.

\begin{figure}[t]
  \centering
    \includegraphics[width=0.4\textwidth]{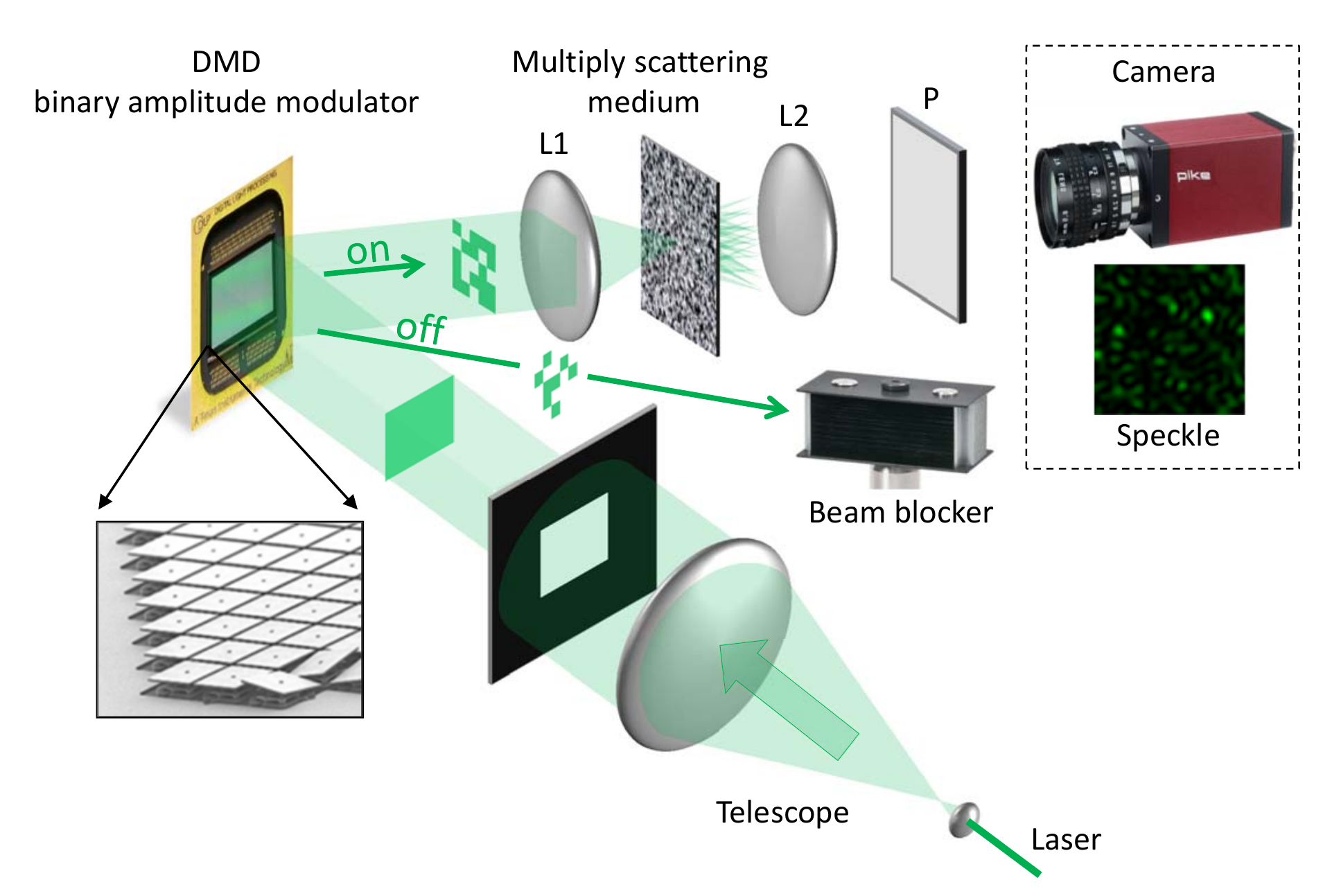}
\caption{Experimental setup of the imager, from \cite{dremeau15}. 
         A monochromatic laser at 532~nm is expanded by a telescope and 
         illuminates an SLM, here, a Texas Instruments DLP9500 DMD with 1920$\times$1080 pixels. 
         The light beam carrying the image is then focused on a random medium by means of 
         a microscope lens. Here, the medium is a thick (several tens of microns) 
         opaque layer of Zinc Oxide nanoparticles deposited on a glass slide. 
         The transmitted light is collected on the far side by a second 
         lens, passes through a polarizer, and is detected by an AVT PIKE F-100 
         monochrome CCD camera. Note that the DMD is only for calibration and 
         display and is not part of the imager itself.}
\label{fig:opticv1}  
\vspace{-0.40cm}
\end{figure}

Here, we investigate the alternative use of a digital micro-mirror device
(DMD) as an SLM, as shown in Fig. \ref{fig:opticv1}. This has many advantages:
DMDs are cheap, fast, and have high pixel counts. However, 
the main drawback of these \textit{binary intensity} modulators is that, without
additional hardware, one can no longer use phase-stepping to measure the complex output field.
Instead of using hardware to measure 
amplitude and phase, we resort to ``phase retrieval'' 
in order to estimate the missing phases from intensity-only 
measurements $|y|^2$.
It should be noted that, in this framework, phase retrieval must be applied
twice successively; first, for the calibration, and second, for the imaging
itself. The success of the second step crucially depends on the first one, as
every error in estimating $\textbf{H}$ results in multiplicative noise (also
called model error) in the imaging step. It should also be noted that
the signal-to-noise ratio is relatively poor, thus we favor Bayesian
phase retrieval techniques where noise may be explicitly modeled.
     
The main contributions of this paper are as follows: 
\begin{itemize}
  \item A new Bayesian phase retrieval algorithm known as phase retrieval Swept AMP (prSAMP). prSAMP originates from prGAMP \cite{schniter15} and SwAMP \cite{manoel14} and is designed to work with noisy ill-conditioned transmission matrices.   
  \item The experimental demonstration that prSAMP is efficient both for calibration of the non-sparse measurement matrix $\textbf{H}$ using binary inputs, and for intensity-only CS imaging of sparse inputs.   
\end{itemize}   
Although our previous studies \cite{liutkus14} demonstrate a proof-of concept that CS-based imaging can be made with multiply scattering materials, we believe that this one-shot imager represents a very significant step toward real-life applications of these techniques.

\section{Theoretical modeling}\label{sec: theoretical_modeling}

Starting from the idealized model of Eq.~\eqref{eq:TM_model}, we formalize the
calibration procedure as in \cite{dremeau15}. Given $P$ known binary input images
of size $N=n_1\times n_2$, $\textbf{X}\in \{0,1\}^{P\times N}$, and their corresponding intensity measurements on
$M$ output pixels, $\textbf{Y}\in \mathbb{R}_+^{M\times P}$, $M$ independent phase retrieval 
problems are solved during the calibration step to estimate the 
transmission matrix $\textbf{H}\in \mathbb{C}^{M\times N}$. Each calibration 
problems is formulated as
\begin{equation}\label{eq:Calib_Model}
\textbf{y}_{\rm m}^T = |\textbf{X} \textbf{h}_{\rm m}^T|,
\end{equation}
where $(\cdot)_m$ indicates the $m$-th row of corresponding matrix and $(\cdot)^T$ is the transpose operator. The
process of recovering a signal from only the magnitude of its 
projections is the goal
of phase retrieval \cite{candes15,iwen15,schniter15}. Apart from
additional noise in the measurements, what makes solving Eq.~\eqref{eq:Calib_Model}
challenging is using binary input patterns, since most well-known phase
retrieval methods work well with complex-valued measurement matrices. We have
fixed this issue by mixing the ideas of Swept Approximate 
Message Passing (SwAMP) \cite{manoel14}, which 
demonstrates good convergence properties over ill-conditioned noisy matrices, 
with the phase retrieval method prGAMP
\cite{schniter15}. The new prSAMP algorithm is explained in the next section.

After calibration, the setup can be used as a generalized CS imager 
with non-linear (intensity) measurements.
In this reconstruction phase, the noiseless model becomes $\textbf{y} = |\textbf{H} \textbf{x}|$.
We use the same prSAMP method, with different priors, to solve both the calibration and reconstruction tasks.

\section{prSAMP algorithm}\label{sec: prSAMP}
In the context of CS, AMP is an iterative algorithm for the reconstruction of a sparse signal from a set of under-determined linear noisy measurements $\textbf{y}=\textbf{H}\textbf{x}+\textbf{w}$, where $\textbf{w}\sim\mathcal{N}(0,\sigma^2)$ \cite{maleki10}. Although this method originates from loopy belief propagation, it does not suffer from the same computational complexity. AMP has been shown to be effective with a minimal number of measurements while being efficient in terms of computational complexity. Using a Bayesian approach, the main loop of AMP consists of iteratively updating the estimated mean $\textbf{x}_{\rm a}$ and variance  $\textbf{x}_{\rm v}$ of the unknown signal until convergence, 
\begin{align}
\textbf{v}^t &= |\textbf{H}|^2 \textbf{x}_{\rm v}^{t-1},\label{eq:amp_v}\\
\bm{\omega}^t &= \textbf{H}\textbf{x}_{\rm a}^{t-1}  
                 - (\textbf{y}-\bm{\omega}^{t-1})\circ\textbf{v}^t\circ(\textbf{v}^{t-1}+\sigma^2)^{-1}, \label{eq:amp_w}\\
\textbf{s}^t &= [|\textbf{H}^*|^2(\textbf{v}^t+\sigma^2)^{-1}]^{-1}, \label{eq:amp_s}\\
\textbf{r}^t &=  \textbf{x}_{\rm a}^t + 
                 \textbf{s}^t\circ
                 \textbf{H}^* [(\textbf{y}-\bm{\omega}^t)\circ(\textbf{v}^t+\sigma^2)^{-1}], \label{eq:amp_r}\\
[\textbf{x}_{\rm a}^t&,  \; \textbf{x}_{\rm v}^t] = \textit{p}_{\rm in}(\textbf{r}^t,\textbf{s}^t), \label{eq:amp_ac}
\end{align}
where $\circ$ is the element-wise Hadamard product, $(\cdot)^{-1}$ is understood to be an element-wise reciprocal,
$(\cdot)^t$ is a time index, $(\cdot)^*$ is the conjugate-transpose,
and $\textit{p}_{\rm in}$ is a function based on the desired signal prior which returns both the mean and variance
estimate of the unknown signal. 
We refer the reader to \cite{krzakala11} for a detailed description of Bayesian AMP.
The calibration and reconstruction phases employ Gaussian and binary priors, respectively \cite{krzakala12,barbier12}. From \cite{popoff10}, we know that the transmission matrices of scattering mediums appear to be i.i.d. random matrices. Therefore, a Gaussian prior for the calibration phase is a reasonable choice. For the reconstruction phase, two binary priors have been investigated based on global (per-image) and local (per-pixel) sparsity, the details of which are
explained in the next section.

Generalized AMP (GAMP) \cite{schniter15} is an extension of AMP for arbitrary
output channels, i.e. $\textbf{y}=q(\textbf{H}\textbf{x}+\textbf{w})$. This adds an
output function, $\textit{p}_{\rm out}$, which is dependent on the stochastic 
description of $q(\cdot)$. In Eqs.~\eqref{eq:amp_w}-\eqref{eq:amp_r}, the terms
$(\textbf{y}-\bm{\omega}^t)\circ(\textbf{v}^{t-1}+\sigma^2)^{-1}$ and
$-(\textbf{v}^t+\sigma^2)^{-1}$ indicate $\textit{p}_{\rm out}$ and $\textit{p}'_{\rm out}$, respectively,
for a Gaussian output channel. One can easily modify these two terms in order to extend the framework to other channels. Following \cite{schniter15}, 
for the phase retrieval problem we have
\begin{align} 
\textit{p}_{\rm out}&=
  \bm{\omega}\circ(\textbf{v}+\sigma^2)^{-1}\circ(r_0\textbf{y}\circ|\bm{\omega}|^{-1}-1), \label{eq:gamp_pout}\\ 
\textit{p}'_{\rm out}&=
  (\textbf{v}+\sigma^2)^{-1}\circ\left[
  \frac{(1-r_0^2)\textbf{y}^2}{(\textbf{v}+\sigma^2)}+\frac{\sigma^2}{\textbf{v}}\right]-\textbf{v}^{-1},\label{eq:gamp_dpout}  
\end{align}
where $r_0=\frac{I_1(\phi)}{I_0(\phi)}$, $I_0$ and $I_1$ are $0^{\rm th}$ and $1^{\rm st}$ order modified Bessel functions of first kind, respectively, and $\bm{\phi}=2\textbf{y}\circ|\bm{\omega}|\circ(\textbf{v}+\sigma^2)^{-1}$. 

The convergence of both AMP
and GAMP has been proved for zero-mean i.i.d. measurement matrices
\cite{bayati11}, however, they do not necessarily converge for generic
matrices \cite{caltagirone14}. There have been some attempts to prevent divergence
of AMP-based methods \cite{vila14,manoel14,cakmak14}. In \cite{manoel14}, the
authors show that a simple change in the main AMP loop may stabilize AMP
significantly. They propose a sequential, or \emph{swept}, random update of the AMP messages
$\textbf{s}^t$, $\textbf{r}^t$, $\textbf{x}_{\rm a}^t$ and $\textbf{x}_{\rm v}^t$,
instead of their standard parallel calculation. By combining the 
swept update ordering and the phase retrieval output channel \eqref{eq:gamp_pout}-\eqref{eq:gamp_dpout} in 
the AMP iteration \eqref{eq:amp_v}-\eqref{eq:amp_ac}, we create a phase retrieval version of SwAMP, denoted as prSAMP,
which we describe in Algorithm 1.     

\begin{figure}
  \centering
    \includegraphics[width=.5\textwidth]{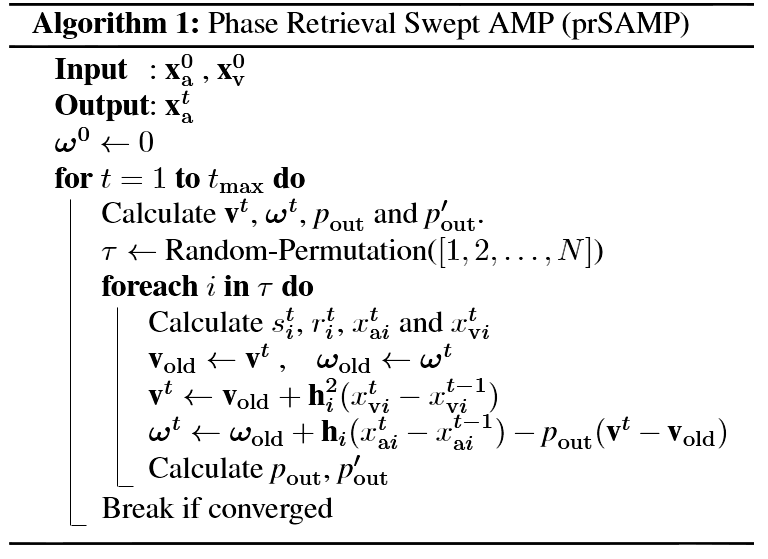}
\end{figure}


\fboxsep=1pt
\fboxrule=0.5pt
\begin{figure}
  \centering
    \fbox{\includegraphics[width=.25\textwidth]{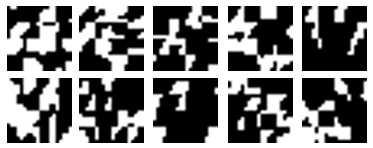}}
\caption{Examples of structured patterns used for calibration.}
\label{fig:calib_patterns}
\vspace{-0.50cm}
\end{figure}

\vspace{-0.50cm}
\section{Experimental Results} 
\label{sec: simulations} 
To investigate the performance of the proposed imaging system, two binary datasets
are constructed from the spatially-sparse MNIST handwritten dataset. The
first, $D_1$, consists of cropped digit images at a resolution of 
$20\times 20$ pixels ($N = 400$). The second, $D_2$, is constructed by 
rescaling the MNIST dataset to $32\times 32$ pixels ($N = 1024$).
Both $D_1$ and $D_2$ retain the original MNIST training/testing 
partition.

For the calibration step, training set $D_1$ is modified 
by randomly exchanging $5\times 5$ blocks of pixels between digit images.
Fig.~\ref{fig:calib_patterns} shows a few samples of these structured patterns.
This structured randomization is done to reduce the effect of correlation
between the DMD pixels.
Additionally, to avoid the possibility of 
completely zero, or very sparse, lines in $\textbf{X}$, see Eq.~\eqref{eq:Calib_Model}, 
we introduce a fixed number of unstructured
i.i.d. Bernoulli random binary patterns to the calibration training set.

\begin{figure}
\centering
  \includegraphics[width=0.22\textwidth]{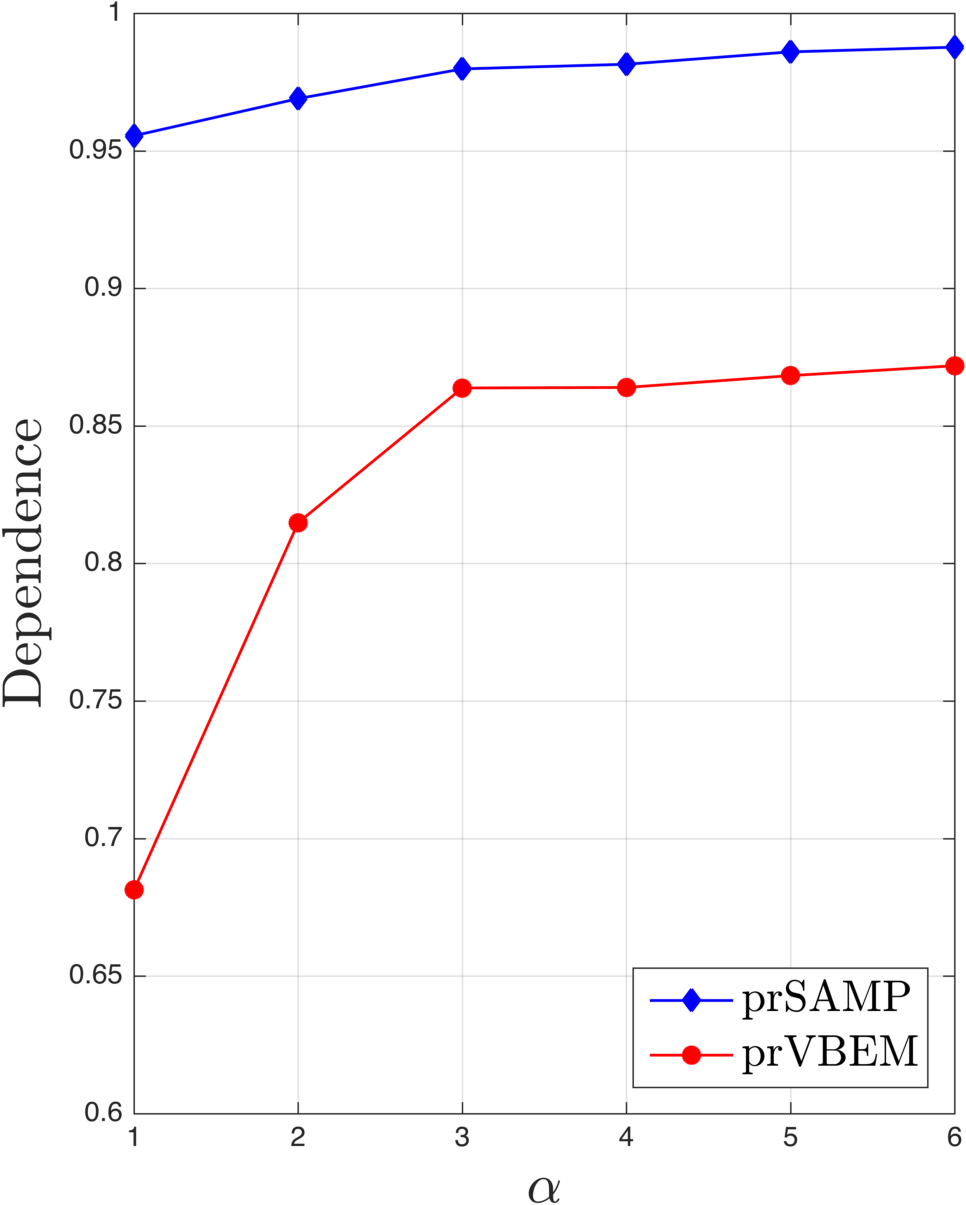}~
  \includegraphics[width=0.22\textwidth]{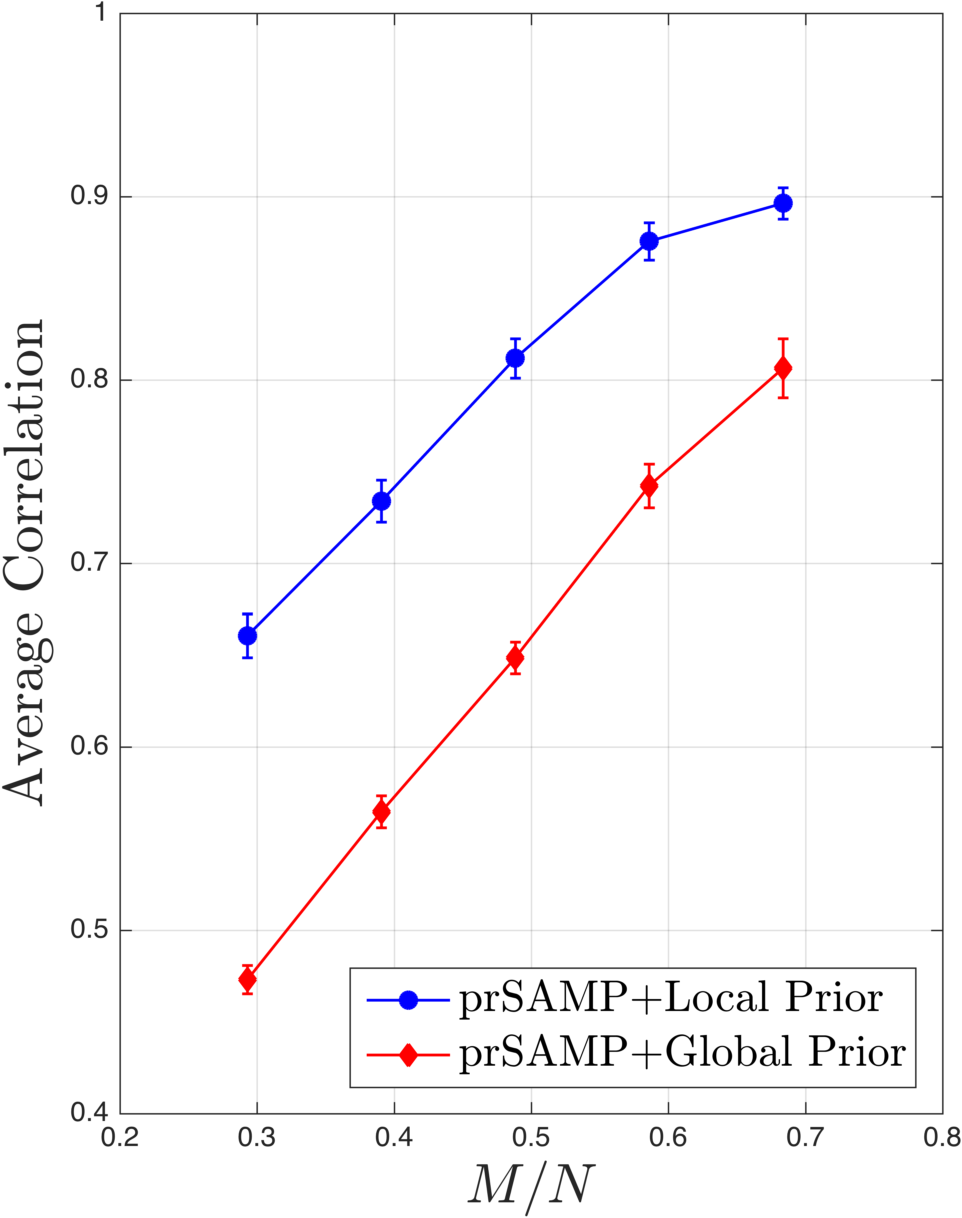}
  
  \caption{   
    \label{fig:calib_rec}
    \textit{Left:}   
      Calibration performance of both prSAMP and prVBEM
      for varying numbers of calibration patterns 
      $P=\alpha N$ which are generated 
      from the $D_1$ dataset.   
    \textit{Right:}   
      Reconstruction performance over 50 digits of $D_2$ ($N = 1024$)
      using
      prSAMP with both global and local binary priors 
      for $M = \left\{300,400,500,600,700\right\}$ output samples.}
\end{figure}

The transmission matrix is then estimated from $P=\alpha N$
calibration images, of which the first
$N$ are Bernoulli random patterns,
for the oversampling ratio $\alpha\geq 1$. At the receptor, $M$ samples are randomly
selected from a $100\times 100$ region of  the output image. 
Fig.~\ref{fig:calib_rec} (left) shows the performance of the proposed calibration 
method for varying values of $\alpha$. In lieu of ground-truth comparisons for
transmission matrix estimation, we assess the calibration performance in
terms of ``dependence,'' the normalized cross-correlation without mean removal, 
$\left\langle  \frac{\textbf{y}}{\norm{\textbf{y}}}, \frac{|\textbf{H}\textbf{x}|}{\norm{\textbf{H}\textbf{x}}} \right\rangle$, 
between observed samples and the predicted output of known input patterns
using the estimated transmission matrix. We measure dependence over 400 digits from the 
testing set of $D_1$. We compare the level of achieved dependence between
prSAMP and prVBEM \cite{dremeau15}, a mean-field variational Bayes
phase retrieval technique we previously employed for the task of transmission matrix calibration
in the context of light focusing.



\fboxsep=1pt
\fboxrule=0.5pt
\begin{figure}
  \centering
    \fbox{\includegraphics[width=.4\textwidth]{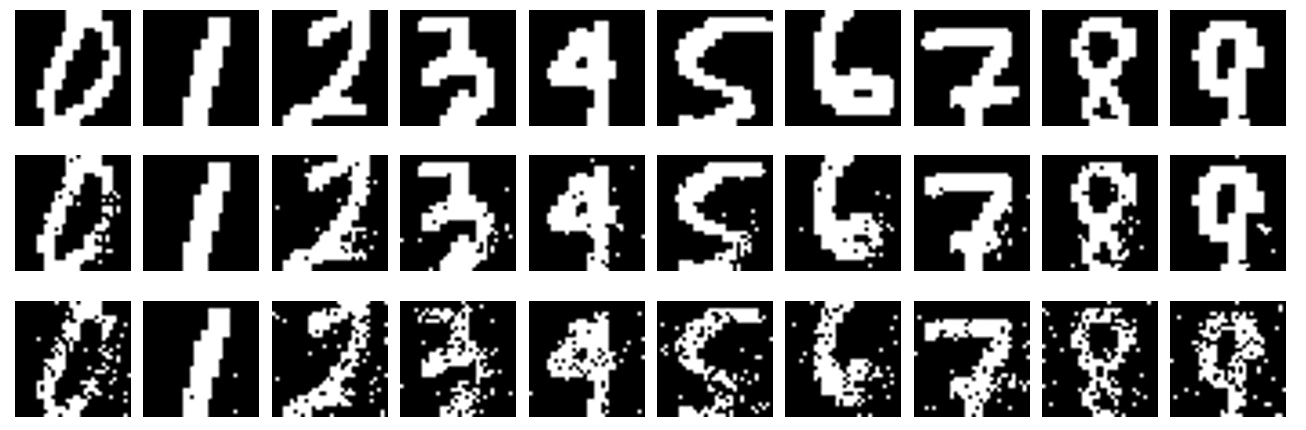}}
\caption{Visual performance of prSAMP reconstruction for 32$\times$32 images at $M=700$. 
         \textit{Top row:} Original images.
         \textit{Middle row:} prSAMP with local prior.
         \textit{Bottom row:} prSAMP with global prior.}
\label{fig:visualMNIST}
\vspace{-0.50cm}
\end{figure}

After calibration, the direct imaging phase can start. As described in
Section  \ref{sec: prSAMP}, the calibration and reconstruction steps are performed
using the same prSAMP algorithm with different input priors.
During calibration, we assume a complex Gaussian prior since the transmission matrix is
modeled as i.i.d. random. However, for reconstruction, a binary
prior is required,
\vspace{-0.5cm}
\begin{align*}\label{eq:binary_prior}
x_{{\rm a}i}^t &= \frac{\rho_i}{z_i} \textit{e}^{\frac{-|1-r_i^t|^2}{2s_i^t}}, \quad
x_{{\rm v}i}^t = x_{{\rm a}i}^t - (x_{{\rm a}i}^t)^2, \numberthis 
\end{align*}

\vspace{-0.35cm}
\noindent where $z_i = (1-\rho_i)\textit{e}^{\frac{-|r_i^t|^2}{2s_i^t}}+\rho_i \textit{e}^{\frac{-|1-r_i^t|^2}{2s_i^t}}$, and
$\rho_i$ indicates the probability of pixel $i$ to be non-zero. 
We use two strategies to set this parameter. The first
is a \textit{global} approach which sets all $\rho_i$ uniformly to the input image
sparsity level which we assume is known up to some tolerance. In the second
\textit{local} approach, we empirically calculate the per-pixel non-zero probability 
using the calibration training set, which is a fast off-line process. 
As the prior calculation must be repeated at each pixel for each sweep of
the prSAMP algorithm, we select the simplest possible prior for the sake of 
computational efficiency. The interested reader may refer to
\cite{tramel15} for a more sophisticated method of using learned priors for
reconstruction tasks.

We next use the $D_2$ dataset to study the effectiveness 
of prSAMP post-calibration reconstruction.
We first perform 
the calibration step to estimate $M=N$ rows of
the transmission matrix using $\alpha=5$, yielding an average calibration
correlation of 97\% over 1024 test digits. For reconstruction, we randomly
choose 50 images from the test set, with five images for each digit.  
The correlation of prSAMP reconstructions to the true inputs, 
using global and local binary priors, are compared in Fig. \ref{fig:calib_rec} (right)
as a function of the measurement rate $M/N$. 
Leveraging the extra information in the local prior provides
an average 14.87\% increase in reconstruction performance over
the global prior.
To visually assess the
quality of recovered images,  Fig.~\ref{fig:visualMNIST} provides one instance
from each digit recovered at $M=700$, with reconstructions using the local and
the global priors. As expected, the
local prior provides better subjective quality with fewer spurious isolated
pixels. 

\section{Conclusion}\label{sec: conclusion} 
In this study, a phase retrieval
compressive imager has been proposed and experimentally evaluated using a
simple optical setup.  The imager has the potential of providing high
resolution images in one shot. We solve the challenging problem of
estimating a complex transmission matrix using binary patterns 
and we solve the phase retrieval problem via swept AMP. 
Finally, we show that we can estimate the transmission matrix accurately,
allowing it to be used for compressive imaging. Further studies 
are necessary to provide faster calibration methods.

\section{Acknowledgment}
The authors would like to thank Christophe Sch{\"u}lke for his
enlightening comments. This research has received funding from the European Research Council under the EU’s 7th Framework Programme (FP/2007- 2013/ERC Grant Agreement 
307087-SPARCS and 278025- COMEDIA) ; and from LABEX WIFI under references 
ANR-10-LABX-24 and ANR-10-IDEX-0001-02-PSL$^\star$.


\bibliography{IntensityImager_arxiv}

\begin{thebibliography}{25}%
\makeatletter
\providecommand \@ifxundefined [1]{%
 \@ifx{#1\undefined}
}%
\providecommand \@ifnum [1]{%
 \ifnum #1\expandafter \@firstoftwo
 \else \expandafter \@secondoftwo
 \fi
}%
\providecommand \@ifx [1]{%
 \ifx #1\expandafter \@firstoftwo
 \else \expandafter \@secondoftwo
 \fi
}%
\providecommand \natexlab [1]{#1}%
\providecommand \enquote  [1]{``#1''}%
\providecommand \bibnamefont  [1]{#1}%
\providecommand \bibfnamefont [1]{#1}%
\providecommand \citenamefont [1]{#1}%
\providecommand \href@noop [0]{\@secondoftwo}%
\providecommand \href [0]{\begingroup \@sanitize@url \@href}%
\providecommand \@href[1]{\@@startlink{#1}\@@href}%
\providecommand \@@href[1]{\endgroup#1\@@endlink}%
\providecommand \@sanitize@url [0]{\catcode `\\12\catcode `\$12\catcode
  `\&12\catcode `\#12\catcode `\^12\catcode `\_12\catcode `\%12\relax}%
\providecommand \@@startlink[1]{}%
\providecommand \@@endlink[0]{}%
\providecommand \url  [0]{\begingroup\@sanitize@url \@url }%
\providecommand \@url [1]{\endgroup\@href {#1}{\urlprefix }}%
\providecommand \urlprefix  [0]{URL }%
\providecommand \Eprint [0]{\href }%
\providecommand \doibase [0]{http://dx.doi.org/}%
\providecommand \selectlanguage [0]{\@gobble}%
\providecommand \bibinfo  [0]{\@secondoftwo}%
\providecommand \bibfield  [0]{\@secondoftwo}%
\providecommand \translation [1]{[#1]}%
\providecommand \BibitemOpen [0]{}%
\providecommand \bibitemStop [0]{}%
\providecommand \bibitemNoStop [0]{.\EOS\space}%
\providecommand \EOS [0]{\spacefactor3000\relax}%
\providecommand \BibitemShut  [1]{\csname bibitem#1\endcsname}%
\let\auto@bib@innerbib\@empty
\bibitem [{\citenamefont {Romberg}(2008)}]{romberg08}%
  \BibitemOpen
  \bibfield  {author} {\bibinfo {author} {\bibfnamefont {J.}~\bibnamefont
  {Romberg}},\ }\href@noop {} {\bibfield  {journal} {\bibinfo  {journal} {IEEE
  Signal Proc. Mag.}\ }\textbf {\bibinfo {volume} {25}},\ \bibinfo {pages} {14}
  (\bibinfo {year} {2008})}\BibitemShut {NoStop}%
\bibitem [{\citenamefont {Eldar}\ and\ \citenamefont
  {Kutyniok}(2012)}]{eldar12}%
  \BibitemOpen
  \bibfield  {author} {\bibinfo {author} {\bibfnamefont {Y.}~\bibnamefont
  {Eldar}}\ and\ \bibinfo {author} {\bibfnamefont {G.}~\bibnamefont
  {Kutyniok}},\ }\href@noop {} {\emph {\bibinfo {title} {Compressed {S}ensing:
  {T}heory and {A}pplications}}}\ (\bibinfo  {publisher} {Cambridge Uni.
  Press},\ \bibinfo {year} {2012})\BibitemShut {NoStop}%
\bibitem [{\citenamefont {Boche}\ \emph {et~al.}(2015)\citenamefont {Boche},
  \citenamefont {Calderbank}, \citenamefont {Kutyniok},\ and\ \citenamefont
  {Vyb{\'\i}ral}}]{boche15}%
  \BibitemOpen
  \bibfield  {author} {\bibinfo {author} {\bibfnamefont {H.}~\bibnamefont
  {Boche}}, \bibinfo {author} {\bibfnamefont {R.}~\bibnamefont {Calderbank}},
  \bibinfo {author} {\bibfnamefont {G.}~\bibnamefont {Kutyniok}}, \ and\
  \bibinfo {author} {\bibfnamefont {J.}~\bibnamefont {Vyb{\'\i}ral}},\
  }\href@noop {} {\emph {\bibinfo {title} {Compressed Sensing and its
  Applications}}}\ (\bibinfo  {publisher} {Springer},\ \bibinfo {year}
  {2015})\BibitemShut {NoStop}%
\bibitem [{\citenamefont {Duarte}\ \emph {et~al.}(2008)\citenamefont {Duarte},
  \citenamefont {Davenport}, \citenamefont {Takhar}, \citenamefont {Laska},
  \citenamefont {Sun}, \citenamefont {Kelly},\ and\ \citenamefont
  {Baraniuk}}]{duarte08}%
  \BibitemOpen
  \bibfield  {author} {\bibinfo {author} {\bibfnamefont {M.}~\bibnamefont
  {Duarte}}, \bibinfo {author} {\bibfnamefont {M.}~\bibnamefont {Davenport}},
  \bibinfo {author} {\bibfnamefont {D.}~\bibnamefont {Takhar}}, \bibinfo
  {author} {\bibfnamefont {J.}~\bibnamefont {Laska}}, \bibinfo {author}
  {\bibfnamefont {T.}~\bibnamefont {Sun}}, \bibinfo {author} {\bibfnamefont
  {K.}~\bibnamefont {Kelly}}, \ and\ \bibinfo {author} {\bibfnamefont
  {R.}~\bibnamefont {Baraniuk}},\ }\href@noop {} {\bibfield  {journal}
  {\bibinfo  {journal} {{IEEE} Signal Proc. Mag.}\ }\textbf {\bibinfo {volume}
  {25}},\ \bibinfo {pages} {83} (\bibinfo {year} {2008})}\BibitemShut {NoStop}%
\bibitem [{\citenamefont {Magalh{\~a}es}\ \emph {et~al.}(2011)\citenamefont
  {Magalh{\~a}es}, \citenamefont {Ara{\'u}jo}, \citenamefont {Correia},
  \citenamefont {Abolbashari},\ and\ \citenamefont {Farahi}}]{magalhaes11}%
  \BibitemOpen
  \bibfield  {author} {\bibinfo {author} {\bibfnamefont {F.}~\bibnamefont
  {Magalh{\~a}es}}, \bibinfo {author} {\bibfnamefont {F.}~\bibnamefont
  {Ara{\'u}jo}}, \bibinfo {author} {\bibfnamefont {M.}~\bibnamefont {Correia}},
  \bibinfo {author} {\bibfnamefont {M.}~\bibnamefont {Abolbashari}}, \ and\
  \bibinfo {author} {\bibfnamefont {F.}~\bibnamefont {Farahi}},\ }\href@noop {}
  {\bibfield  {journal} {\bibinfo  {journal} {Appl. Opt.}\ }\textbf {\bibinfo
  {volume} {50}},\ \bibinfo {pages} {405} (\bibinfo {year} {2011})}\BibitemShut
  {NoStop}%
\bibitem [{\citenamefont {Shrekenhamer}\ \emph {et~al.}(2013)\citenamefont
  {Shrekenhamer}, \citenamefont {Watts},\ and\ \citenamefont
  {Padilla}}]{shrekenhamer13}%
  \BibitemOpen
  \bibfield  {author} {\bibinfo {author} {\bibfnamefont {D.}~\bibnamefont
  {Shrekenhamer}}, \bibinfo {author} {\bibfnamefont {C.}~\bibnamefont {Watts}},
  \ and\ \bibinfo {author} {\bibfnamefont {W.}~\bibnamefont {Padilla}},\
  }\href@noop {} {\bibfield  {journal} {\bibinfo  {journal} {Opt. Express}\
  }\textbf {\bibinfo {volume} {21}},\ \bibinfo {pages} {12507} (\bibinfo {year}
  {2013})}\BibitemShut {NoStop}%
\bibitem [{\citenamefont {Fergus}\ \emph {et~al.}(2006)\citenamefont {Fergus},
  \citenamefont {Torralba},\ and\ \citenamefont {Freeman}}]{fergus06}%
  \BibitemOpen
  \bibfield  {author} {\bibinfo {author} {\bibfnamefont {R.}~\bibnamefont
  {Fergus}}, \bibinfo {author} {\bibfnamefont {A.}~\bibnamefont {Torralba}}, \
  and\ \bibinfo {author} {\bibfnamefont {W.}~\bibnamefont {Freeman}},\
  }\href@noop {} {\bibfield  {journal} {\bibinfo  {journal} {MIT CSAIL Tech.
  Report}\ } (\bibinfo {year} {2006})}\BibitemShut {NoStop}%
\bibitem [{\citenamefont {Zhao}\ \emph {et~al.}(2012)\citenamefont {Zhao},
  \citenamefont {Gong}, \citenamefont {Chen}, \citenamefont {Li}, \citenamefont
  {Wang}, \citenamefont {Xu},\ and\ \citenamefont {Han}}]{zhao12}%
  \BibitemOpen
  \bibfield  {author} {\bibinfo {author} {\bibfnamefont {C.}~\bibnamefont
  {Zhao}}, \bibinfo {author} {\bibfnamefont {W.}~\bibnamefont {Gong}}, \bibinfo
  {author} {\bibfnamefont {M.}~\bibnamefont {Chen}}, \bibinfo {author}
  {\bibfnamefont {E.}~\bibnamefont {Li}}, \bibinfo {author} {\bibfnamefont
  {H.}~\bibnamefont {Wang}}, \bibinfo {author} {\bibfnamefont {W.}~\bibnamefont
  {Xu}}, \ and\ \bibinfo {author} {\bibfnamefont {S.}~\bibnamefont {Han}},\
  }\href@noop {} {\bibfield  {journal} {\bibinfo  {journal} {Appl. Phys.
  Lett.}\ }\textbf {\bibinfo {volume} {101}},\ \bibinfo {pages} {141123}
  (\bibinfo {year} {2012})}\BibitemShut {NoStop}%
\bibitem [{\citenamefont {Dr{\'e}meau}\ \emph {et~al.}(2015)\citenamefont
  {Dr{\'e}meau}, \citenamefont {Liutkus}, \citenamefont {Martina},
  \citenamefont {Katz}, \citenamefont {Sch{\"u}lke}, \citenamefont {Krzakala},
  \citenamefont {Gigan},\ and\ \citenamefont {Daudet}}]{dremeau15}%
  \BibitemOpen
  \bibfield  {author} {\bibinfo {author} {\bibfnamefont {A.}~\bibnamefont
  {Dr{\'e}meau}}, \bibinfo {author} {\bibfnamefont {A.}~\bibnamefont
  {Liutkus}}, \bibinfo {author} {\bibfnamefont {D.}~\bibnamefont {Martina}},
  \bibinfo {author} {\bibfnamefont {O.}~\bibnamefont {Katz}}, \bibinfo {author}
  {\bibfnamefont {C.}~\bibnamefont {Sch{\"u}lke}}, \bibinfo {author}
  {\bibfnamefont {F.}~\bibnamefont {Krzakala}}, \bibinfo {author}
  {\bibfnamefont {S.}~\bibnamefont {Gigan}}, \ and\ \bibinfo {author}
  {\bibfnamefont {L.}~\bibnamefont {Daudet}},\ }\href@noop {} {\bibfield
  {journal} {\bibinfo  {journal} {Opt. Express}\ }\textbf {\bibinfo {volume}
  {23}},\ \bibinfo {pages} {11898} (\bibinfo {year} {2015})}\BibitemShut
  {NoStop}%
\bibitem [{\citenamefont {Gao}\ \emph {et~al.}(2014)\citenamefont {Gao},
  \citenamefont {Liang}, \citenamefont {Li},\ and\ \citenamefont
  {Wang}}]{gao14}%
  \BibitemOpen
  \bibfield  {author} {\bibinfo {author} {\bibfnamefont {L.}~\bibnamefont
  {Gao}}, \bibinfo {author} {\bibfnamefont {J.}~\bibnamefont {Liang}}, \bibinfo
  {author} {\bibfnamefont {C.}~\bibnamefont {Li}}, \ and\ \bibinfo {author}
  {\bibfnamefont {L.}~\bibnamefont {Wang}},\ }\href@noop {} {\bibfield
  {journal} {\bibinfo  {journal} {Nature}\ }\textbf {\bibinfo {volume} {516}}
  (\bibinfo {year} {2014})}\BibitemShut {NoStop}%
\bibitem [{\citenamefont {Popoff}\ \emph {et~al.}(2010)\citenamefont {Popoff},
  \citenamefont {Lerosey}, \citenamefont {Carminati}, \citenamefont {Fink},
  \citenamefont {Boccara},\ and\ \citenamefont {Gigan}}]{popoff10}%
  \BibitemOpen
  \bibfield  {author} {\bibinfo {author} {\bibfnamefont {S.}~\bibnamefont
  {Popoff}}, \bibinfo {author} {\bibfnamefont {G.}~\bibnamefont {Lerosey}},
  \bibinfo {author} {\bibfnamefont {R.}~\bibnamefont {Carminati}}, \bibinfo
  {author} {\bibfnamefont {M.}~\bibnamefont {Fink}}, \bibinfo {author}
  {\bibfnamefont {A.}~\bibnamefont {Boccara}}, \ and\ \bibinfo {author}
  {\bibfnamefont {S.}~\bibnamefont {Gigan}},\ }\href@noop {} {\bibfield
  {journal} {\bibinfo  {journal} {Phys. Rev. Lett}\ }\textbf {\bibinfo {volume}
  {104}},\ \bibinfo {pages} {100601} (\bibinfo {year} {2010})}\BibitemShut
  {NoStop}%
\bibitem [{\citenamefont {Liutkus}\ \emph {et~al.}(2014)\citenamefont
  {Liutkus}, \citenamefont {Martina}, \citenamefont {Popoff}, \citenamefont
  {Chardon}, \citenamefont {Katz}, \citenamefont {Lerosey}, \citenamefont
  {Gigan}, \citenamefont {Daudet},\ and\ \citenamefont {Carron}}]{liutkus14}%
  \BibitemOpen
  \bibfield  {author} {\bibinfo {author} {\bibfnamefont {A.}~\bibnamefont
  {Liutkus}}, \bibinfo {author} {\bibfnamefont {D.}~\bibnamefont {Martina}},
  \bibinfo {author} {\bibfnamefont {S.}~\bibnamefont {Popoff}}, \bibinfo
  {author} {\bibfnamefont {G.}~\bibnamefont {Chardon}}, \bibinfo {author}
  {\bibfnamefont {O.}~\bibnamefont {Katz}}, \bibinfo {author} {\bibfnamefont
  {G.}~\bibnamefont {Lerosey}}, \bibinfo {author} {\bibfnamefont
  {S.}~\bibnamefont {Gigan}}, \bibinfo {author} {\bibfnamefont
  {L.}~\bibnamefont {Daudet}}, \ and\ \bibinfo {author} {\bibfnamefont
  {I.}~\bibnamefont {Carron}},\ }\href@noop {} {\bibfield  {journal} {\bibinfo
  {journal} {Sci. Rep.}\ }\textbf {\bibinfo {volume} {4}} (\bibinfo {year}
  {2014})}\BibitemShut {NoStop}%
\bibitem [{\citenamefont {Schniter}\ and\ \citenamefont
  {Rangan}(2015)}]{schniter15}%
  \BibitemOpen
  \bibfield  {author} {\bibinfo {author} {\bibfnamefont {P.}~\bibnamefont
  {Schniter}}\ and\ \bibinfo {author} {\bibfnamefont {S.}~\bibnamefont
  {Rangan}},\ }\href@noop {} {\bibfield  {journal} {\bibinfo  {journal} {{IEEE}
  Trans. on Signal Proc.}\ }\textbf {\bibinfo {volume} {63}},\ \bibinfo {pages}
  {1043} (\bibinfo {year} {2015})}\BibitemShut {NoStop}%
\bibitem [{\citenamefont {Manoel}\ \emph {et~al.}(2014)\citenamefont {Manoel},
  \citenamefont {Krzakala}, \citenamefont {Tramel},\ and\ \citenamefont
  {Zdeborov{\'a}}}]{manoel14}%
  \BibitemOpen
  \bibfield  {author} {\bibinfo {author} {\bibfnamefont {A.}~\bibnamefont
  {Manoel}}, \bibinfo {author} {\bibfnamefont {F.}~\bibnamefont {Krzakala}},
  \bibinfo {author} {\bibfnamefont {E.}~\bibnamefont {Tramel}}, \ and\ \bibinfo
  {author} {\bibfnamefont {L.}~\bibnamefont {Zdeborov{\'a}}},\ }in\ \href@noop
  {} {\emph {\bibinfo {booktitle} {Int. Conf. on Machine Learning}}}\ (\bibinfo
  {year} {2014})\BibitemShut {NoStop}%
\bibitem [{\citenamefont {Candes}\ \emph {et~al.}(2015)\citenamefont {Candes},
  \citenamefont {Eldar}, \citenamefont {Strohmer},\ and\ \citenamefont
  {Voroninski}}]{candes15}%
  \BibitemOpen
  \bibfield  {author} {\bibinfo {author} {\bibfnamefont {E.}~\bibnamefont
  {Candes}}, \bibinfo {author} {\bibfnamefont {Y.}~\bibnamefont {Eldar}},
  \bibinfo {author} {\bibfnamefont {T.}~\bibnamefont {Strohmer}}, \ and\
  \bibinfo {author} {\bibfnamefont {V.}~\bibnamefont {Voroninski}},\
  }\href@noop {} {\bibfield  {journal} {\bibinfo  {journal} {SIAM Rev.}\
  }\textbf {\bibinfo {volume} {57}},\ \bibinfo {pages} {225} (\bibinfo {year}
  {2015})}\BibitemShut {NoStop}%
\bibitem [{\citenamefont {Iwen}\ \emph {et~al.}(2015)\citenamefont {Iwen},
  \citenamefont {Viswanathan},\ and\ \citenamefont {Wang}}]{iwen15}%
  \BibitemOpen
  \bibfield  {author} {\bibinfo {author} {\bibfnamefont {M.}~\bibnamefont
  {Iwen}}, \bibinfo {author} {\bibfnamefont {A.}~\bibnamefont {Viswanathan}}, \
  and\ \bibinfo {author} {\bibfnamefont {Y.}~\bibnamefont {Wang}},\ }\href@noop
  {} {\bibfield  {journal} {\bibinfo  {journal} {arXiv preprint:1501.02377}\ }
  (\bibinfo {year} {2015})}\BibitemShut {NoStop}%
\bibitem [{\citenamefont {Maleki}(2010)}]{maleki10}%
  \BibitemOpen
  \bibfield  {author} {\bibinfo {author} {\bibfnamefont {A.}~\bibnamefont
  {Maleki}},\ }\emph {\bibinfo {title} {Approximate Message Passing Algorithms
  for Compressed Sensing}},\ \href@noop {} {Ph.D. thesis},\ \bibinfo  {school}
  {Stanford Uni.} (\bibinfo {year} {2010})\BibitemShut {NoStop}%
\bibitem [{\citenamefont {Krzakala}\ \emph {et~al.}(2011)\citenamefont
  {Krzakala}, \citenamefont {M{\'e}zard}, \citenamefont {Sausset},
  \citenamefont {Sun},\ and\ \citenamefont {Zdeborov{\'a}}}]{krzakala11}%
  \BibitemOpen
  \bibfield  {author} {\bibinfo {author} {\bibfnamefont {F.}~\bibnamefont
  {Krzakala}}, \bibinfo {author} {\bibfnamefont {M.}~\bibnamefont
  {M{\'e}zard}}, \bibinfo {author} {\bibfnamefont {F.}~\bibnamefont {Sausset}},
  \bibinfo {author} {\bibfnamefont {Y.}~\bibnamefont {Sun}}, \ and\ \bibinfo
  {author} {\bibfnamefont {L.}~\bibnamefont {Zdeborov{\'a}}},\ }\href@noop {}
  {\bibfield  {journal} {\bibinfo  {journal} {Phys. Rev. X}\ }\textbf {\bibinfo
  {volume} {2}} (\bibinfo {year} {2011})}\BibitemShut {NoStop}%
\bibitem [{\citenamefont {Krzakala}\ \emph {et~al.}(2012)\citenamefont
  {Krzakala}, \citenamefont {M{\'e}zard}, \citenamefont {Sausset},
  \citenamefont {Sun},\ and\ \citenamefont {Zdeborov{\'a}}}]{krzakala12}%
  \BibitemOpen
  \bibfield  {author} {\bibinfo {author} {\bibfnamefont {F.}~\bibnamefont
  {Krzakala}}, \bibinfo {author} {\bibfnamefont {M.}~\bibnamefont
  {M{\'e}zard}}, \bibinfo {author} {\bibfnamefont {F.}~\bibnamefont {Sausset}},
  \bibinfo {author} {\bibfnamefont {Y.}~\bibnamefont {Sun}}, \ and\ \bibinfo
  {author} {\bibfnamefont {L.}~\bibnamefont {Zdeborov{\'a}}},\ }\href@noop {}
  {\bibfield  {journal} {\bibinfo  {journal} {J. Stat. Mech.: Theory \& Exp.}\
  }\textbf {\bibinfo {volume} {2012}},\ \bibinfo {pages} {P08009} (\bibinfo
  {year} {2012})}\BibitemShut {NoStop}%
\bibitem [{\citenamefont {Barbier}\ \emph {et~al.}(2012)\citenamefont
  {Barbier}, \citenamefont {Krzakala}, \citenamefont {M{\'e}zard},\ and\
  \citenamefont {Zdeborov{\'a}}}]{barbier12}%
  \BibitemOpen
  \bibfield  {author} {\bibinfo {author} {\bibfnamefont {J.}~\bibnamefont
  {Barbier}}, \bibinfo {author} {\bibfnamefont {F.}~\bibnamefont {Krzakala}},
  \bibinfo {author} {\bibfnamefont {M.}~\bibnamefont {M{\'e}zard}}, \ and\
  \bibinfo {author} {\bibfnamefont {L.}~\bibnamefont {Zdeborov{\'a}}},\ }in\
  \href@noop {} {\emph {\bibinfo {booktitle} {{IEEE} Allerton Conf. on Comm.,
  Control, \& Computing}}}\ (\bibinfo {year} {2012})\ pp.\ \bibinfo {pages}
  {800--807}\BibitemShut {NoStop}%
\bibitem [{\citenamefont {Bayati}\ and\ \citenamefont
  {Montanari}(2011)}]{bayati11}%
  \BibitemOpen
  \bibfield  {author} {\bibinfo {author} {\bibfnamefont {M.}~\bibnamefont
  {Bayati}}\ and\ \bibinfo {author} {\bibfnamefont {A.}~\bibnamefont
  {Montanari}},\ }\href@noop {} {\bibfield  {journal} {\bibinfo  {journal}
  {{IEEE} Trans. on Info. Theory}\ }\textbf {\bibinfo {volume} {57}},\ \bibinfo
  {pages} {764} (\bibinfo {year} {2011})}\BibitemShut {NoStop}%
\bibitem [{\citenamefont {Caltagirone}\ \emph {et~al.}(2014)\citenamefont
  {Caltagirone}, \citenamefont {Zdeborov{\'a}},\ and\ \citenamefont
  {Krzakala}}]{caltagirone14}%
  \BibitemOpen
  \bibfield  {author} {\bibinfo {author} {\bibfnamefont {F.}~\bibnamefont
  {Caltagirone}}, \bibinfo {author} {\bibfnamefont {L.}~\bibnamefont
  {Zdeborov{\'a}}}, \ and\ \bibinfo {author} {\bibfnamefont {F.}~\bibnamefont
  {Krzakala}},\ }in\ \href@noop {} {\emph {\bibinfo {booktitle} {{IEEE} Int.
  Symp. on Info. Theory}}}\ (\bibinfo {year} {2014})\ pp.\ \bibinfo {pages}
  {1812--1816}\BibitemShut {NoStop}%
\bibitem [{\citenamefont {Vila}\ \emph {et~al.}(2015)\citenamefont {Vila},
  \citenamefont {Schniter}, \citenamefont {Rangan}, \citenamefont {Krzakala},\
  and\ \citenamefont {Zdeborov{\'a}}}]{vila14}%
  \BibitemOpen
  \bibfield  {author} {\bibinfo {author} {\bibfnamefont {J.}~\bibnamefont
  {Vila}}, \bibinfo {author} {\bibfnamefont {P.}~\bibnamefont {Schniter}},
  \bibinfo {author} {\bibfnamefont {S.}~\bibnamefont {Rangan}}, \bibinfo
  {author} {\bibfnamefont {F.}~\bibnamefont {Krzakala}}, \ and\ \bibinfo
  {author} {\bibfnamefont {L.}~\bibnamefont {Zdeborov{\'a}}},\ }in\ \href@noop
  {} {\emph {\bibinfo {booktitle} {{IEEE} Int. Conf. on Acoustics, Speech, \&
  Signal Proc.}}}\ (\bibinfo {year} {2015})\ pp.\ \bibinfo {pages}
  {2021--2025}\BibitemShut {NoStop}%
\bibitem [{\citenamefont {{\c C}akmak}\ \emph {et~al.}(2014)\citenamefont {{\c
  C}akmak}, \citenamefont {Winther},\ and\ \citenamefont {Fleury}}]{cakmak14}%
  \BibitemOpen
  \bibfield  {author} {\bibinfo {author} {\bibfnamefont {B.}~\bibnamefont {{\c
  C}akmak}}, \bibinfo {author} {\bibfnamefont {O.}~\bibnamefont {Winther}}, \
  and\ \bibinfo {author} {\bibfnamefont {B.}~\bibnamefont {Fleury}},\ }in\
  \href@noop {} {\emph {\bibinfo {booktitle} {{IEEE} Info. Theory Work.}}}\
  (\bibinfo {year} {2014})\ pp.\ \bibinfo {pages} {192--196}\BibitemShut
  {NoStop}%
\bibitem [{\citenamefont {Tramel}\ \emph {et~al.}(2016)\citenamefont {Tramel},
  \citenamefont {Dr{\'e}meau},\ and\ \citenamefont {Krzakala}}]{tramel15}%
  \BibitemOpen
  \bibfield  {author} {\bibinfo {author} {\bibfnamefont {E.}~\bibnamefont
  {Tramel}}, \bibinfo {author} {\bibfnamefont {A.}~\bibnamefont {Dr{\'e}meau}},
  \ and\ \bibinfo {author} {\bibfnamefont {F.}~\bibnamefont {Krzakala}},\
  }\href@noop {} {\bibfield  {journal} {\bibinfo  {journal} {J. Stat. Mech.:
  Theory \& Exp.}\ } (\bibinfo {year} {2016})},\ \bibinfo {note} {to
  appear}\BibitemShut {NoStop}%
\end{thebibliography}%
\end{document}